\documentclass[conference]{IEEEtran}
\IEEEoverridecommandlockouts
\usepackage{cite}
\usepackage{amsmath,amssymb,amsfonts}
\usepackage{algorithmic}
\usepackage{textcomp}
\usepackage{xcolor}
\usepackage{hyperref}
\usepackage{url}
\usepackage{graphicx}
\usepackage{enumitem}
\usepackage{multirow}
\usepackage{tabularx}
\usepackage{bbm}
\usepackage{booktabs}
\usepackage{tablefootnote}
\usepackage[T1]{fontenc}
\def\BibTeX{{\rm B\kern-.05em{\sc i\kern-.025em b}\kern-.08em
    T\kern-.1667em\lower.7ex\hbox{E}\kern-.125emX}}
\begin{document}

\title{Learn while Unlearn: An Iterative Unlearning Framework for Generative Language Models
\thanks{$^{\bf \dagger}$Equal Contribution, $^{\bf *}$Corresponding Author.}
}

\author{
{\bf Haoyu Tang$^{1,\dagger}$}, {\bf Ye Liu$^{1,\dagger}$}, {\bf Xi Zhao$^{2}$}, {\bf Xukai Liu$^{1}$}, {\bf Yanghai Zhang$^{1}$}\\ {\bf Kai Zhang$^{1,*}$},  {\bf Xiaofang Zhou$^{2}$}, {\bf Enhong Chen$^{1}$} \\
        $^1$ University of Science and Technology of China,\\
        $^2$ The Hong Kong University of Science and Technology\\
        \{haoyu\_t, chthollylxk, yhzhang0612\}@mail.ustc.edu.cn, yeliu.liuyeah@gmail.com\\
        \{kkzhang08, cheneh\}@ustc.edu.cn, \{xzhaoca, zxf\}@cse.ust.hk\\
}


\maketitle

\begin{abstract}
Recent advances in machine learning, particularly in Natural Language Processing (NLP), have produced powerful models trained on vast datasets. However, these models risk leaking sensitive information, raising privacy concerns. In response, regulatory measures such as the European Union's General Data Protection Regulation (GDPR) have driven increasing interest in Machine Unlearning techniques, which enable models to selectively forget specific data entries. Early unlearning approaches primarily relied on pre-processing methods, while more recent research has shifted towards training-based solutions. Despite their effectiveness, a key limitation persists: most methods require access to original training data, which is often unavailable. Additionally, directly applying unlearning techniques bears the cost of undermining the model's expressive capabilities. To address these challenges, we introduce the \textbf{I}terative \textbf{C}ontrastive \textbf{U}nlearning (\textbf{ICU}) framework, which consists of three core components: A Knowledge Unlearning Induction module designed to target specific knowledge for removal using an unlearning loss; A Contrastive Learning Enhancement module to preserve the model's expressive capabilities against the pure unlearning goal; And an Iterative Unlearning Refinement module that  dynamically adjusts the unlearning process through ongoing evaluation and updates. Experimental results demonstrate the efficacy of our ICU method in unlearning sensitive information while maintaining the model's overall performance, offering a promising solution for privacy-conscious machine learning applications.

\end{abstract}


\section{Introduction}

The rapid advancement of deep learning, particularly in Natural Language Processing (NLP), has led to the widespread deployment of powerful generative language models (GLMs) such as GPT-4~\cite{achiam2023gpt}, Claude 3~\cite{TheC3}, and Google Gemini~\cite{team2023gemini}. These models exhibit remarkable capabilities in text generation, summarization, and conversational AI, transforming numerous real-world applications. However, this progress has also raised critical concerns regarding data privacy, security, and compliance with evolving regulations. One of the most pressing challenges is the risk of unintentional memorization and leakage of sensitive information, such as personally identifiable data, proprietary content, and confidential records contained in training datasets~\cite{carlini2021extracting,brown2022does,carlini2022quantifying}.

To mitigate these risks, regulatory frameworks like the General Data Protection Regulation (GDPR)~\cite{voigt2017eu} have introduced principles such as the “Right To Be Forgotten” (RTBF)~\cite{mantelero2013eu, villaronga2018humans}, which necessitate the ability to remove or selectively forget specific information from trained machine learning models. This has driven extensive research into machine unlearning, a field that aims to ensure models can eliminate particular knowledge while maintaining overall functionality and performance~\cite{shi2024muse, hu2025blur}.

\begin{figure}[t]
  \centering
  \includegraphics[width=0.49\textwidth]{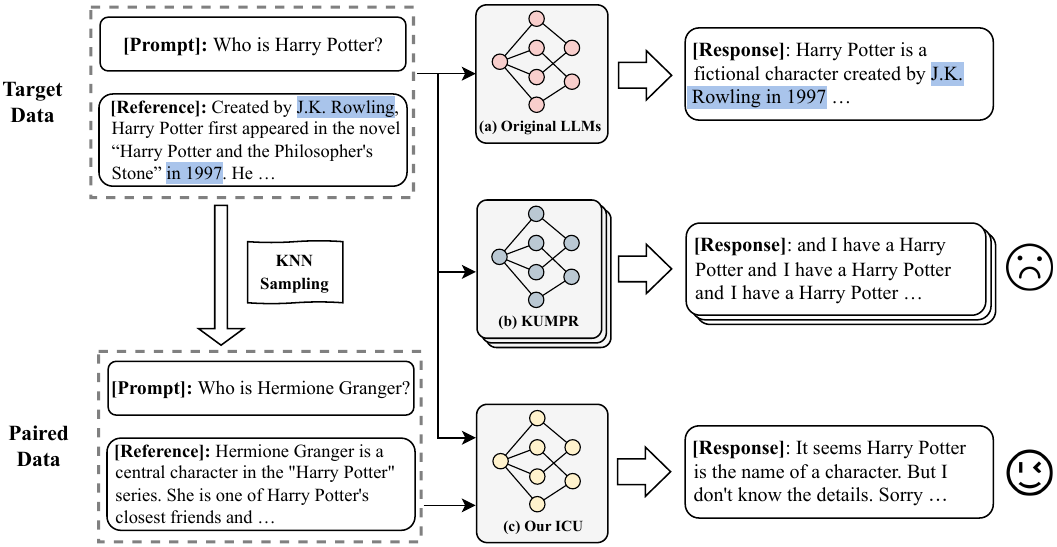}
  \caption{Difference among the generated sequences of (a) original model, (b) model unlearned by KUMPR~\cite{jang2023knowledge} and (c) model unlearned by our method.}
  \vspace{-5mm}
  \label{fig:intro}
\end{figure}

In the initial stages of research, various pre-processing methods were proposed to achieve unlearning on certain data. For instance, Kandpal et al.\cite{kandpal2022deduplicating} found that the likelihood of Generative Language Models regenerating training sequences is correlated with the frequency of those sequences in the training set. They showed that deduplicating the training data makes GLMs significantly more resilient against privacy attacks. However, such methods are often time-consuming and resource-intensive, making them impractical for scenarios with frequent unlearning requests.

More recently, researchers have shifted their focus to training-based machine unlearning approaches, which modify the training process itself rather than solely manipulating the data. For example, SISA~\cite{bourtoule2021machine} partitions the original dataset into several non-overlapping shards and then aggregates models trained on these separate shards. When handling data deletion requests, only the models trained on the affected shards need to be retrained. KGA~\cite{wang2023kga} introduces an additional dataset, using it to fine-tune the original model alongside the original dataset. Nonetheless, both of these methods assume that the training data remain accessible during the unlearning process. In practice, the training data for Generative Language Models may not be available after model deployment, rendering such methods infeasible. A recent method, Knowledge Unlearning for Mitigating Privacy Risks (KUMPR)~\cite{jang2023knowledge}, was designed to address this scenario.\footnote{We refer to this method as KUMPR as the authors did not provide a specific name.} KUMPR reverses the training objective by maximizing the negative log-likelihood for target tokens and uses metrics to determine whether the model has ``forgotten'' a target sequence.

However, directly reversing the training objective of GLMs can go beyond merely forgetting the target knowledge, leading to the loss of their overall expressive capabilities. As illustrated in Figure~\ref{fig:intro}, given a training sample with the prefix ``\textit{Who is Harry Potter?}'' and its reference suffix, the goal is to make the model ``forget'' specific knowledge, such as ``\textit{J.K. Rowling}'' and ``\textit{1997}''. Although existing methods like KUMPR successfully forget the key information contained in the reference suffix, the resulting model loses its general expressive ability, often generating repetitive, nonsensical text. For example, in Figure~\ref{fig:intro} (b), the model repeats ``and I have a Harry Potter'', indicating that it has collapsed. To address this limitation, we propose an Iterative Contrastive Unlearning (ICU) framework, which aims to achieve knowledge unlearning while preserving the model’s overall generalization ability.

More specifically, our ICU framework comprises three components: (1) A Knowledge Unlearning Induction (KUI) module, which applies an unlearning loss to remove the specific knowledge effectively; (2) A Contrastive Learning Enhancement (CLE) module, which samples paired data from the analogous documents of target knowledge. Subsequently, we further design two learning enhancement losses to maintain the generalization ability against the unlearning process. (3) An Iterative Unlearning Refinement (IUR) module, which assess the unlearning extent on specific data pieces and updates the unlearning dataset dynamically, effectively preventing over-unlearning and excessive performance degradation.
Finally, experimental results and analyses from various perspectives demonstrate that our proposed method outperforms baseline approaches in terms of balancing maintaining performance and improving unlearning efficiency.

In brief, our contributions are as follows:
\begin{itemize}[leftmargin=0.1cm]
    \item We conduct an in-depth study on the unlearning techniques for Generative Language Models, specifically focusing on maintaining the expression ability while achieving effectively unlearning, an area that has been largely overlooked by previous researchers.
    \item We propose the ICU framework, which consists of three components: Knowledge Unlearning Induction, Contrastive Learning Enhancement, and Iterative Unlearning Refinement.
    \item We perform extensive experiments on three different backbone models of varying sizes, demonstrating the effectiveness of our proposed method. Our code is available at \href{https://github.com/himalalps/ICU}{https://github.com/himalalps/ICU}.
\end{itemize}


\section{Related work}
The related work can be categorized into two parts, including: (1) Machine Unlearning and (2) Generative Language Models.

\subsection{Machine unlearning}

Machine unlearning, introduced by Cao et al.~\cite{cao2015towards}, aims to protect machine learning models from extraction attacks by removing specific data in such a way that the model behaves as if the data were never part of the training set. Traditional approaches~\cite{nguyen2022survey} that exclude specific data from training datasets and retrain the model are highly time-consuming and resource-intensive, making them impractical for modern deep neural networks. Similar methods involving retraining~\cite{bourtoule2021machine, kumar2023privacy} also struggle with scalability, particularly when handling numerous deletion requests or when comprehensive datasets are not readily available. To address these challenges, researchers have explored approximate unlearning techniques~\cite{golatkar2020eternal, mehta2022deep}. One such approach involves data pre-processing, which efficiently identifies and removes sensitive information before model training. Kandpal et al.\cite{kandpal2022deduplicating} applied this method to structured private data, such as phone numbers and medical records, and found it effective. However, challenges arise when dealing with unstructured data, as pre-processing may not fully remove all sensitive information~\cite{brown2020language} and cannot comprehensively address ongoing deletion demands~\cite{brown2022does}.

Recent studies~\cite{jang2023knowledge, kassem2023preserving, yao2023large, eldan2023s,zhang2024negative} have focused on fine-tuning Generative Language Models to tackle machine unlearning challenges. Jang et al.\cite{jang2023knowledge} proposed a novel approach by reversing the traditional training objective, aiming to maximize rather than minimize the negative log-likelihood of tokens designated for forgetting. Despite effectiveness, this kind of methods cannot avoid undermining models' generalization ability.
Other recent methods~\cite{chen2023unlearn, gao2024ethos, gu2024second, kassem2023preserving, wang2023kga,maini2024tofu,yao2023large,li2024wmdp} use diverse techniques such as knowledge gap alignment and reinforcement learning. Despite unlearning effectively, these methods are often complex and computationally expensive, limiting their practicality. For instance, Gu et al.\cite{gu2024second} utilized second-order information (Hessian) to provide stronger guarantees for data removal while maintaining model utility, but this approach requires substantial computational resources for Hessian approximation, making it difficult to play a role in real scenarios. Pawelczyk et al.\cite{pawelczyk2023context} applied in-context methods in unlearning approaches, yet not effective for generation tasks.

\begin{figure*}[t]
  \centering
  \includegraphics[width=0.8\textwidth]{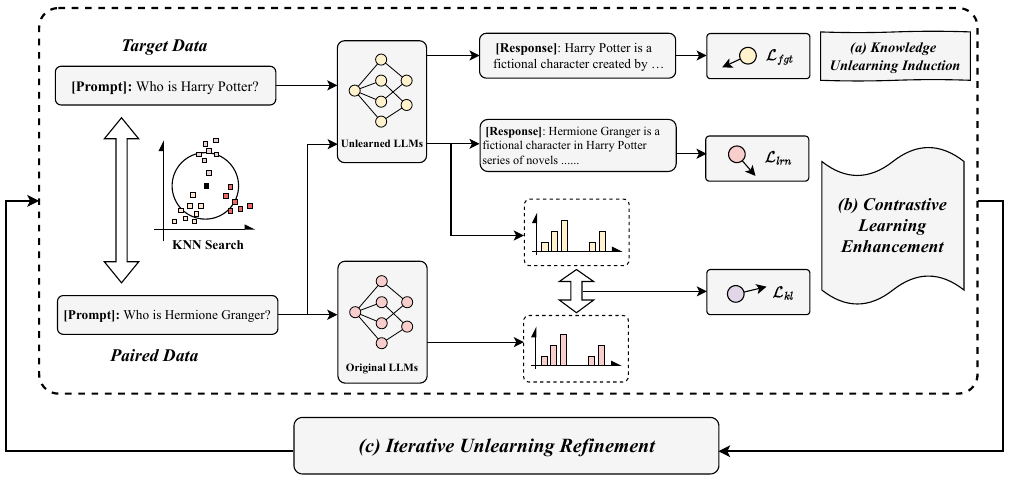}
  \caption{The structure of Iterative Contrastive Unlearning framework. It consists of three parts: (a)~Knowledge Unlearning Induction (KUI), (b)~Contrastive Learning Enhancement (CLE), and (c)~Iterative Unlearning Refinement (IUR).}
  \vspace{-5mm}
  \label{fig:model}
\end{figure*}

\subsection{Generative language models}
Generative Language Models are designed to understand, generate, and predict human language~\cite{zhao2023survey}, which have gained considerable attention in recent years.

Traditional language models, such as rule-based approaches~\cite{weizenbaum1966eliza} and statistical models~\cite{brown1992class}, generate outputs that resemble human language but do not perfectly reflect the training data. Early neural network models in NLP, including Recurrent Neural Networks (RNNs)~\cite{werbos1990backpropagation}, faced limitations due to their sequential processing architecture, which resulted in high computational demands and hindered scalability. 

The introduction of the transformer architecture~\cite{vaswani2017attention} revolutionized NLP by enabling the effective capture of contextual relationships through self-attention mechanisms~\cite{liu2023enhancing}. Decoder-only models, including GPT-4~\cite{achiam2023gpt}, Claude 3~\cite{TheC3}, and others~\cite{anil2023palm, team2023gemini, touvron2023llama}, have demonstrated exceptional performance across a wide range of NLP tasks~\cite{liu2024empowering,liu2024onenet}. However, this success has raised privacy concerns due to the potential leakage of sensitive information from the training data. Additionally, increasing model capacity has led to greater demands for training data and computational resources~\cite{achiam2023gpt, TheC3, hoffmann2022training, ouyang2022training}, creating significant challenges for researchers working on machine unlearning for advanced models.

\section{Problem statement}
Formally speaking, the machine unlearning task can be formulated as follows:

Given the target data and model $\{D_{\mathrm{fgt}}, f_\theta\}$, where $D_{\text{fgt}} = \{\boldsymbol{x}_{i}^{\mathrm{fgt}}\}^{M}_{i=1}$ is the collection of $M$ pieces of data to be forgotten and $f_\theta$ is the original model with its parameters denoted as $\theta$, machine unlearning aims to modify the parameters $\theta$ such that the retention of previously learned information about $D_{\mathrm{fgt}}$ is minimized while maintaining desirable model performance and meeting specified constraints.

\label{sec:ICU_framework}
\section{ICU framework}

\subsection{Model overview}

We propose a novel Iterative Contrastive Unlearning (ICU) framework, illustrated in Figure~\ref{fig:model}. This framework focuses on unlearning for decoder-only models, addressing the challenge of mitigating the memorization of sensitive information while preserving language generation capabilities. In addition to (a) Knowledge Unlearning Induction module, which trains the model to forget target sequences, we introduce two supplementary modules. (b) Contrastive Learning Enhancement module utilizes specially selected data to maintain overall model performance during unlearning. Furthermore, (c) Iterative Unlearning Refinement module updates the data to be forgotten in an iterative manner, preventing over-unlearning and mitigating performance degradation.

\subsection{Knowledge Unlearning Induction}

For a sample in the forget set $\boldsymbol{x}^{\mathrm{fgt}}\in D_{\mathrm{fgt}}$, the sequence of tokens is denoted as $\boldsymbol{x}=(x_1,x_2,\ldots,x_T)$. Following the approach of Jang et al.\cite{jang2023knowledge}, we \textit{negate} the original negative log-likelihood of the target token sequences to induce the model to forget these sequences. The unlearning objective is defined as:
\begin{equation}
    \mathcal{L}_{\mathrm{fgt}}=\sum_{t=t_0}^{T}\log P_\theta(x_t^{\mathrm{fgt}}|x_{<t}^{\mathrm{fgt}}),\label{eq:fgt}
\end{equation}
where $x_{<t} = (x_1,x_2,\ldots,x_{t-1})$ denotes the first $t$ tokens of the sequence, $t_0$ is the length of tokens provided to the model, and $P_\theta(x_t|x_{<t})$ represents the conditional probability of predicting the next token $x_t$ given the previous tokens $x_{<t}$, with $\theta$ representing the model parameters.

\label{sec:CLE}
\subsection{Contrastive Learning Enhancement}

To maintain the model's stable expression capabilities during unlearning, we propose training the model simultaneously on analogous data. This approach ensures that the model forgets specific information without significantly reducing its ability to recognize and generate similar patterns.

\subsubsection{Analogous data construction.}
The first step is to construct a data pool related to the forget set $D_{\mathrm{fgt}}$, aiming to identify data samples that contain similar but different knowledge. Specifically, we retrieve documents from \textit{Wiki} that belong to the same category as the forget set $D_{\mathrm{fgt}}$ but contain different key concepts. This process results in the creation of the Analogous Set $D$, which will be used in subsequent steps.

\subsubsection{KNN sampling.}
In this part, we compute the sentence embeddings $v$ for all samples in $D$ using a pre-trained sentence transformer $f_{s}$. For each sample $\boldsymbol{x}^{\mathrm{fgt}}$ in $D_{\mathrm{fgt}}$ with its embedding $v_x^{\mathrm{fgt}}=f_{s}(\boldsymbol{x}^{\mathrm{fgt}})$, we employ K-Nearest Neighbors (KNN)~\cite{douze2024faiss} to identify the nearest ($K=1$) embedding $\hat{v_x}$ and the corresponding sample $\boldsymbol{\hat{x}}$. All $\boldsymbol{\hat{x}}$ are then collected to form $D_{\mathrm{lrn}}$:
\begin{equation}
    \boldsymbol{\hat{x}}=\underset{\boldsymbol{x}\in D\setminus D_{\mathrm{fgt}}}{\mathrm{argmin}}\ dis(v_x,\hat{v_x}),
\end{equation}
where $dis(\cdot)$ is the cosine similarity function used in KNN search.

\subsubsection{Learning enhancement.}
The retrieved paired data sample for $\boldsymbol{x}^{\mathrm{fgt}}$ is denoted as $\boldsymbol{\hat{x}}=\boldsymbol{x}^{\mathrm{lrn}}\in D_{\mathrm{lrn}}$. In contrast to the unlearning objective, we force the model to learn patterns from these paired token sequences using the negative log-likelihood, defined as follows:
\begin{equation}
    \mathcal{L}_{\mathrm{lrn}}=-\sum_{t=t_0}^{T}\log P_\theta(x_t^{\mathrm{lrn}}|x_{<t}^{\mathrm{lrn}}),\label{eq:lrn}
\end{equation}

Additionally, we apply Kullback-Leibler (KL) divergence~\cite{kullback1951kullback} to guide the model to approximate the original model's distribution for data intended to be retained, following Yao et al.\cite{yao2023large}:
\begin{equation}
    \mathcal{L}_{\mathrm{kl}}=\sum_{t=t_0}^{T}KL[P_{\theta_0}(x_t^{\mathrm{lrn}}|x_{<t}^{\mathrm{lrn}})||P_{\theta}(x_t^{\mathrm{lrn}}|x_{<t}^{\mathrm{lrn}})],\label{eq:kl}
\end{equation}
where $\theta_0$ denotes the parameters of the original model.

Here, $\mathcal{L}_{\mathrm{lrn}}$ and $\mathcal{L}_{\mathrm{kl}}$ seems similar but they actually differ. $\mathcal{L}_{\mathrm{lrn}}$ ensures the model generates paired data instead of the original target data, while $\mathcal{L}_{\mathrm{kl}}$ preserves the expression ability of the original model. For experiments with only loss \eqref{eq:lrn} or \eqref{eq:kl}, please refer to the ablation study results in Figure~\ref{fig:ablation}. 

\label{sec:IUR}
\subsection{Iterative Unlearning Refinement}

Unlike conventional machine learning techniques, validating the efficacy of unlearning presents challenges in identifying a suitable validation set, as $D_{\mathrm{fgt}}$ is integrated into the training phase. Thus, establishing an appropriate stopping criterion is essential. After each training epoch, the model's performance relative to the target data is evaluated using the metrics described below: 
\begin{equation}
\begin{aligned}
    \mathrm{BERT Score}(\boldsymbol{x}, f_\theta(x_{<t_0}))<a, \\
    \mathrm{BLEU}(\boldsymbol{x},f_\theta(x_{<t_0}))<b,\label{eq:IUR}
\end{aligned}
\end{equation}
where $\boldsymbol{x}$ is the referenced target sample, $f_\theta(x_{<t_0})$ denotes the output of the model provided input sequence $x_{<t_0}$, and $a$ and $b$ are predefined thresholds for the iteration process. We empirically determine that a specific token sequence $\boldsymbol{x}$ is considered ``forgotten'' if \eqref{eq:IUR} is satisfied. Samples deemed ``forgotten'' are excluded from subsequent epochs. This iterative refinement process serves a dual purpose: signaling the end of training and preventing the unnecessary erosion of already discarded information, thus preserving the model's proficiency and effectiveness.

\textbf{Bilingual Evaluation Understudy (BLEU)}~\cite{papineni2002bleu} is a metric originally used to evaluate machine translation quality by measuring the similarity between a model-generated token sequence and one or more reference translations, based on n-gram comparisons. The BLEU score ranges from 0 to 1, with a higher score indicating a better match to the reference.

\textbf{BERTScore}~\cite{zhang2019bertscore} leverages contextual embeddings from BERT~\cite{devlin2018bert} models to assess the similarity between two provided sentences. Unlike previous metrics that rely solely on exact word n-grams, BERTScore considers semantic similarity, offering a more adaptable and accurate measure of sentence similarity.

\subsection{Training}

The objectives in \eqref{eq:fgt}, \eqref{eq:lrn} and \eqref{eq:kl} are jointly used to optimize the model during unlearning. The training process is governed by minimizing the following loss function:
\begin{equation}
    \mathcal{L}=\mathcal{L}_{\mathrm{fgt}}+\alpha\mathcal{L}_{\mathrm{lrn}}+\beta\mathcal{L}_{\mathrm{kl}},\label{eq:loss}
\end{equation}
where $\alpha,\beta>0$ are positive hyper-parameters that control the contributions of the different optimizing objectives. 

\section{Experiments}

In this section, we first describe the datasets used for training and evaluation, as well as the metrics employed to assess performance. Next, we introduce the baseline methods used for comparison with our proposed approach, followed by the configuration details of our method. Finally, we present and analyze the experimental results.

\label{sec:datasets}
\subsection{Datasets} 

To evaluate ICU's learning and unlearning capabilities, we selected two types of datasets: Target Datasets, which assess the unlearning performance, and Downstream Dataset, which evaluates the original capabilities of the models.

\paragraph{Target Dataset.} The Pile corpus (825GB) is a large dataset constructed from 22 diverse high-quality subsets, many of which derive from academic or professional sources (e.g. books, open source code)~\cite{gao2020pile}. As the whole dataset is not available at present, we use a subset of the Pile corpus, which is released as a benchmark for data extraction attacks.\footnote{https://github.com/google-research/lm-extraction-benchmark} Designed to be easy-to-extract, the subset contains 15,000 samples, randomly sampled from the Pile training dataset. Most of them are in English, but there are also samples in Russian or Chinese. Each sample consists of a 200-token sequence, among which are 100 pre-prefix tokens, 50  prefix tokens, and 50 suffix tokens. Following Jang et al.\cite{jang2023knowledge}, we only use the prefix and suffix tokens thus $t_0$ is set to 50. This choice of benchmark follows standard practices like Jang et al.\cite{jang2023knowledge} and Kassem et al.\cite{kassem2023preserving}, where the same benchmark is widely used to evaluate unlearning methods. While alternative benchmarks such as WMDP~\cite{li2024wmdp} exist, they represent different datasets designed for the same unlearning task, and thus do not fundamentally differ in the nature of the evaluation. Since all these benchmarks focus on generative unlearning tasks, additional experiments on alternative datasets are unnecessary, as they would likely yield similar insights without providing significant new information.

\paragraph{Downstream Dataset.} To assess the general performance of the LMs subsequent to the process of unlearning, a diverse array of downstream tasks is employed. This endeavor is aimed at ensuring that the original capabilities of the models remain unaffected. 

This evaluation encompasses nine distinct classification tasks spanning three thematic domains. Specifically, these domains include linguistic reasoning tasks such as Hellaswag~\cite{zellers2019hellaswag} and Lambada~\cite{paperno2016lambada}, as well as assessments of commonsense reasoning through Winogrande~\cite{sakaguchi2021winogrande} and COPA~\cite{gordon2012COPA}. Additionally, scientific reasoning abilities are evaluated through tasks such as ARC-Easy~\cite{clark2018arc}, ARC-Challenge~\cite{clark2018arc}, Piqa~\cite{bisk2020piqa}, MathQA~\cite{amini2019mathqa}, and PubmedQA~\cite{jin2019pubmedqa}. 

Furthermore, four dialogue tasks, namely Wizard of Wikipedia~\cite{dinan2018wizardwikipedia}, Empathetic Dialogues~\cite{rashkin2018empathetic}, Blended Skill Talk~\cite{smith2020can}, and Wizard of Internet~\cite{komeili2021internet}, are used to gauge the model's proficiency in generating coherent responses. In addition, we measure the perplexity of the unlearned models on the validation set of Pile and Wikitext. Following \cite{jang2023knowledge}, we use the test set for Lambada and the validation set for the remaining tasks.

\label{metric}
\subsection{Metrics} 

As stated in Section~\ref{sec:datasets}, we assess both learning and unlearning capabilities using two types of datasets. For evaluating unlearning performance, we follow Jang et al.\cite{jang2023knowledge} and examine the forgetting effect on the target unlearning data using Extraction Likelihood (EL) and Memorization Accuracy (MA). In the work of Jang et al.\cite{jang2023knowledge}, these metrics are also used as stopping criteria during training, making them unsuitable as sole evaluation metrics. As mentioned in Section~\ref{sec:IUR}, we additionally employ BERTScore and BLEU to measure forgetting during the Iterative Unlearning Refinement (IUR) module.

To evaluate the original capabilities of unlearned models, we first use the basic metrics provided in the downstream datasets, obtaining \textit{Accuracy} for classification task and \textit{F1} for dialogue tasks. Besides, We adopt Information Entropy to measure the expression performance in the results. Furthermore, we also employ GPT-4 and Human Evaluation to assess the text generated by the unlearned models. The following details the metrics used:

\textbf{Extraction Likelihood (EL)} is introduced by Jang et al.\cite{jang2023knowledge} to  measure the average success rate of varying extraction attacks quantified via getting the n-gram overlap of generated and target token sequences. It is computed by the following equation:
\begin{align}
    \mathrm{EL}_n(\boldsymbol{x})=\frac{\sum_{t=1}^{T-n}\mathrm{OVERLAP}_n(f_\theta(x_{<t}),x_{\geq t})}{T-n},\\
    \mathrm{OVERLAP}_n(\boldsymbol{a},\boldsymbol{b})=\frac{\sum_{c\in ng(\boldsymbol{a})}\mathbbm{1}\{c\in ng(\boldsymbol{b})\}}{|ng(\boldsymbol{a})|}.
\end{align}

\textbf{Memorization Accuracy (MA)}~\cite{tirumala2022memorization} quantifies how much model $f_\theta$ has memorized the given token sequences, which is defined as follows:
\begin{equation}
    \mathrm{MA}(\boldsymbol{x})=\frac{\sum_{t=1}^{T-1}\mathbbm{1}\{\mathrm{argmax}(P_\theta(\cdot|x_{<t})=x_t\}}{T-1}.
\end{equation}

\textbf{Information Entropy}
quantifies the average uncertainty in a set of outcomes, reflecting the amount of information produced by a random source~\cite{vajapeyam2014understanding}. Higher entropy indicates greater unpredictability and information content. 
Mathematically, entropy ($H$) is defined for a discrete random variable $X$ with possible outcomes $\{x_1,x_2,\ldots,x_n\}$ and corresponding probabilities $\{p_1,p_2,\ldots,p_n\}$ as:
\begin{equation}
    H(x)=-\sum_{i=1}^np_i\log_2p_i.
\end{equation}

\textbf{GPT Evaluation} uses GPT-4~\cite{achiam2023gpt} to evaluate the unlearned models in two perspectives: whether the model generates text without prior knowledge of key information in the referenced target data and whether the generated sequences are coherent. The prompts can be found in our publicly available codes.

\textbf{Human Evaluation} indicates that we hire human experts to determine the goodness of the generated response, following the goal in the \textbf{GPT Evaluation} part. 

\textbf{Normalized Scores} provide a balanced score across all metrics related to unlearning and generative capabilities of different methods:
\begin{equation}
\text{s}=\sum_{\uparrow}M_i+\sum_{\downarrow}M_i  
\end{equation}
where $\sum_{\uparrow} M_i$ is the sum of normalized metrics that should be maximized (e.g., Entropy, Acc, F1, GPT) and $\sum_{\downarrow} M_i$ is the sum of metrics that should be minimized (e.g., EL, MA, BERTScore, PPL). Inspired by \cite{de2023choice}, We compute $s_z$ and $s_m$, which uses z-score and min-max normalization respectively.

\subsection{Baseline methods} 

Our experiments use the \textsc{GPT-Neo} model family (125M, 1.3B, 2.7B)~\cite{black2021gpt}, which is pre-trained on the Pile corpus. Following Jang et al.\cite{jang2023knowledge}, we utilize the \textsc{Opt} model family (125M, 1.3B, 2.7B)~\cite{zhang2022opt}, which is pre-trained on a deduplicated version of the Pile as well as other corpus, serving as our baseline method for deduplication since the deduplicated version of \textsc{GPT-Neo} by Kandpal et al.\cite{kandpal2022deduplicating} is not publicly available. For the approximate unlearning methods, we include KUMPR~\cite{jang2023knowledge}, DPO~\cite{maini2024tofu}, KL~\cite{maini2024tofu} and LLMU~\cite{yao2023large} as other baseline methods on \textsc{GPT-Neo} models to show the effectiveness of our proposed method. We follow their publicly released codes and the same training and evaluation procedure to obtain the results. For experiment on model families other than \textsc{GPT-Neo}, we also include results on TinyLlama 1.1B model~\cite{zhang2024tinyllama} in Section~\ref{app:generalization} to demonstrate our model's generalization ability.

\begin{table*}[t]
  \caption{Results showing the average of five random samples. Cls Avg. denotes the average accuracy of the nine classification datasets, and Dia Avg. denotes the average F1 score of the four dialogue datasets. The best comparable performances of unlearning are \textbf{bolded} and second best \underline{underlined}.}
  \label{tab:main}
  \centering
  \setlength{\tabcolsep}{6pt} 
  \renewcommand{\arraystretch}{1} 
  \begin{tabular}{l|c|ccc|ccc|cc|c|cc}
    \toprule
    \multirow{2}{*}{\textbf{Model}} & \textbf{\#} & \textbf{EL}$_{10}$ & \textbf{MA} & \textbf{BERT} & \textbf{Entropy} & \textbf{Cls Avg.} & \textbf{Dia Avg.} & \textbf{Pile} & \textbf{Wikitext} & \textbf{GPT} & $\mathbf{s_z}$ & $\mathbf{s_m}$ \\
         & \textbf{Params} & (\%) $\downarrow$ & (\%) $\downarrow$ & (F1) $\downarrow$ & $\uparrow$ & (ACC) $\uparrow$ & (F1) $\uparrow$ & (PPL) $\downarrow$ & (PPL) $\downarrow$ & $\uparrow$ & $\uparrow$ & $\uparrow$\\
    \midrule
    \midrule
    \textsc{Neo} (before unlearning) &  & 51.9 & 76.8 & 70.3 & 4.139 & 43.5 & 10.0 & 20.1 & 38.0 & - & - & - \\
    \midrule
    \textsc{Opt} & \multirow{6}{*}{125M} & 7.5 & 52.9 & 49.2 & 3.014 & \underline{42.7} & \textbf{10.8} & 29.1 & \textbf{38.0} & 3.08 & 2.23 & 1.31 \\
    \textsc{Neo} + KUMPR & & \textbf{0.7} & \textbf{19.1} & \textbf{29.7} & 0.712 & 35.1 & 3.7 & >1000 & >1000 & 1.05 & -5.72 & -1.79 \\
    \textsc{Neo} + DPO & & 17.4  & \underline{49.4} & \underline{42.2} & 1.643 & 39.0 & 1.8 & 61.2 & 158.5 & 1.99 & -5.01 & -1.34 \\
    \textsc{Neo} + KL &  & 4.9 & 56.2 & 54.3 & \underline{3.670} & 42.6 & 9.9 & 27.0 & 54.0 & \underline{3.64} & \underline{2.69} & \underline{1.45} \\
    \textsc{Neo} + LLMU &  & \underline{3.3} & 58.7 & 43.1 & 1.825 & \underline{42.7} & \underline{10.3} & \underline{23.6} & 47.4 & 2.98 & 2.03 & 1.19 \\
    \textbf{\textsc{Neo} + ICU (ours)} & & 4.4 & 55.6 & 53.3 & \textbf{3.833} & \textbf{43.3} & \underline{10.3} & \textbf{21.6} & \underline{40.1} & \textbf{3.92} & \textbf{3.78} & \textbf{1.84} \\
    \midrule
    \midrule
    \textsc{Neo} (before unlearning) & & 98.2 & 92.3 & 86.3 & 4.640 & 49.7 & 12.3 & 13.2 & 18.7 & - & - & - \\
    \midrule
    \textsc{Opt} & \multirow{6}{*}{1.3B} & 31.0 & 67.8 & 65.8 & \underline{3.856} & \textbf{51.7} & \textbf{13.3} & 18.0 & \textbf{19.2} & \underline{3.63} & 0.87 & 0.75 \\
    \textsc{Neo} + KUMPR & & \textbf{0.8} & \textbf{8.1} & \textbf{26.6} & 0.817 & 34.0 & 0.1 & >1000 & >1000 & 1.26 & -6.61 & -2.00 \\
    \textsc{Neo} + DPO &  & 20.0 & 58.4 & 55.1 & 2.564 & 44.8 & 4.8 & 26.6 & 44.6 & 2.56 & -2.09 & -0.29 \\
    \textsc{Neo} + KL &  & \underline{3.7} & 61.5 & \underline{38.7} & 1.627 & 47.4 & 11.8 & 17.5 & 26.0 & 2.18 & 1.31 & 0.90 \\
    \textsc{Neo} + LLMU &  & 4.4 & 63.3 & 41.4 & 1.805 & 47.4 & 11.9 & \underline{17.5} & 25.7 & 2.43 & \underline{1.38} & \underline{0.92} \\
    \textbf{\textsc{Neo} + ICU (ours)} & & 4.7 & \underline{51.3} & 52.7 & \textbf{3.900} & \underline{49.0} & \underline{12.1} & \textbf{14.1} & \underline{19.3} & \textbf{4.33} & \textbf{5.15} & \textbf{2.24} \\
    \midrule
    \midrule
    \textsc{Neo} (before unlearning) & & 96.7 & 93.7 & 90.2 & 4.719 & 52.4 & 12.3 & 12.0 & 16.2 & - & - & - \\
    \midrule
    \textsc{Opt} & \multirow{6}{*}{2.7B} & 34.4 & 70.1 & 66.8 & \textbf{3.921} & \textbf{53.9} & \textbf{13.7} & 16.3 & \textbf{16.7} & \underline{3.65} & 0.51 & 0.78 \\
    \textsc{Neo} + KUMPR & & \textbf{1.4} & \textbf{18.7} & \textbf{26.5} & 0.519 & 34.0 & 5.4 & >1000 & >1000 & 1.00 & -7.01 & -2.00 \\
    \textsc{Neo} + DPO &  & 20.4 & 58.6 & 57.5 & 2.772 & 49.0 & 7.6 & 24.1 & 36.5 & 2.66 & -1.63 & 0.02 \\
    \textsc{Neo} + KL &  & \underline{3.4} & 62.5 & 41.3 & 1.730 & 51.3 & \underline{12.5} & \underline{14.7} & 20.9 & 2.39 & \underline{1.72} & \underline{1.20} \\
    \textsc{Neo} + LLMU &  & 4.9 & 63.2 & \underline{40.7} & 1.639 & 50.9 & 12.4 & 16.0 & 22.4 & 2.26 & 1.30 & 1.06 \\
    \textbf{\textsc{Neo} + ICU (ours)} & & 4.5 & \underline{48.3} & 52.7 & \underline{3.725} & \underline{52.1} & 12.1 & \textbf{13.1} & \underline{17.0} & \textbf{4.40} & \textbf{5.11} & \textbf{2.34} \\
    \bottomrule
  \end{tabular}
  \vspace{-5mm}
\end{table*}

\subsection{Configurations} 

For each model size (125M, 1.3B, 2.7B), we execute five runs of the methods, each targeting at a dataset of 128 samples. In the Contrastive Learning Enhancement module (Section~\ref{sec:CLE}), we utilize the remaining Pile subset for the Analogous Data Construction, and all-MiniLM-L6-v2 model to conduct KNN sampling. 
The model is optimized by Adam~\cite{kingma2014adam} with a learning rate of $5e-6$, and $\alpha=0.5,\beta=1.0$. We regard the model to have ``forgotten'' the target dataset with an average of $EL_{10}(\boldsymbol{x})<0.0499$ and $MA(\boldsymbol{x})<0.5994$ following Jang et al.\cite{jang2023knowledge}. The filtering thresholds during iteration are $a=0.3$ and $b=0.01$ as introduced in Section~\ref{sec:IUR}. Specifically, for BERTScore, only 2\% of outputs fell below $0.3$, while the average score exceeded $0.7$, with many instances close to $1$. For BLEU, the largest value below $0.01$ was $1e-78$, and the smallest above $0.01$ was $0.0159$, making $0.01$ a clear dividing point. These thresholds effectively balance precision and applicability based on the observed data distributions.

We run all the experiments on a Linux server with one 2.60GHz Intel Xeon Platinum 8358 CPU and NVIDIA GeForce RTX 3090 GPUs. We use one GPU for 125M models with batch size of 8. With Deepspeed Stage 2, we use three GPUs for 1.3B and six GPUs for 2.7B respectively with batch size of 4.

\label{sec:main_results}
\subsection{Main results}

The results of all methods are summarized in Table~\ref{tab:main}. Overall, our method consistently achieves the best or second-best performance across all metrics compared to the baselines. Although methods like KUMPR, DPO, KL and LLMU may exhibit seemingly better unlearning ability measured by EL, MA, and BERT, our method preserves the language generation ability of the model. Compared to \textsc{Opt}, our model unlearns the original model better. Overall, our method achieves the best normalized scores across all model sizes, showing the best trade-off between unlearning and performance. Additionally, we observe several interesting phenomena:

\begin{figure}[t]
  \centering
  \includegraphics[width=0.45\textwidth]{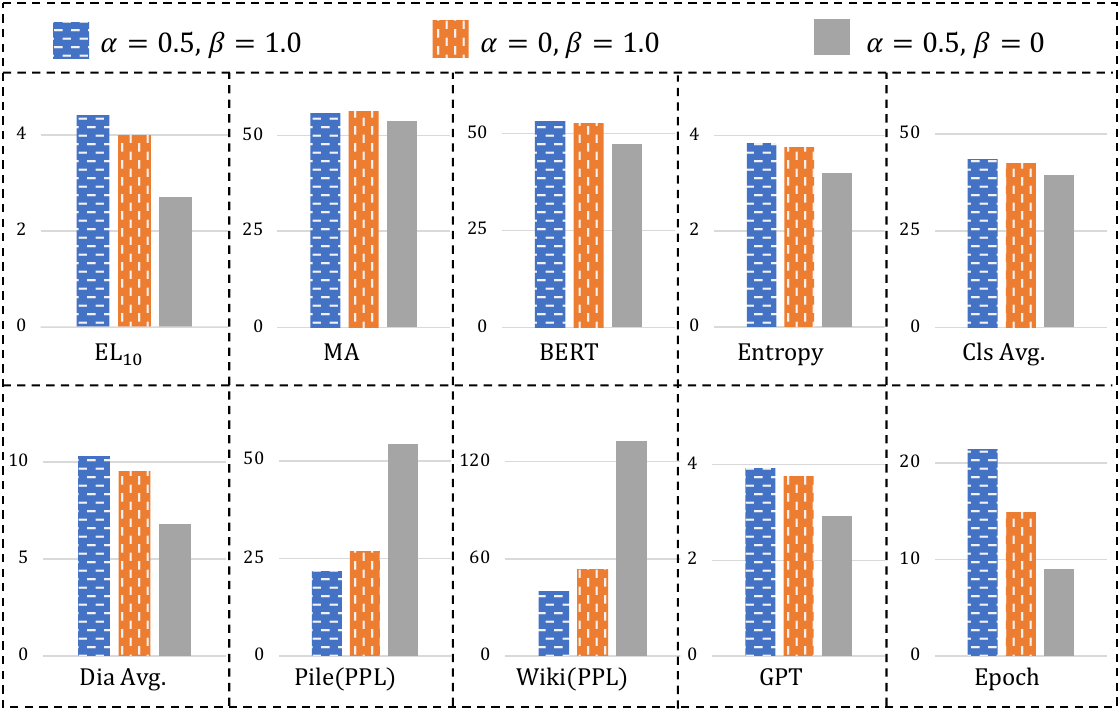}
  \caption{Ablation results on hyperparameters ($\alpha,\beta$) for \textsc{GPT-Neo} 125M.}
  \vspace{-5mm}
  \label{fig:ablation}
\end{figure}

\begin{figure*}[t]
  \centering
  \includegraphics[width=0.9\textwidth]{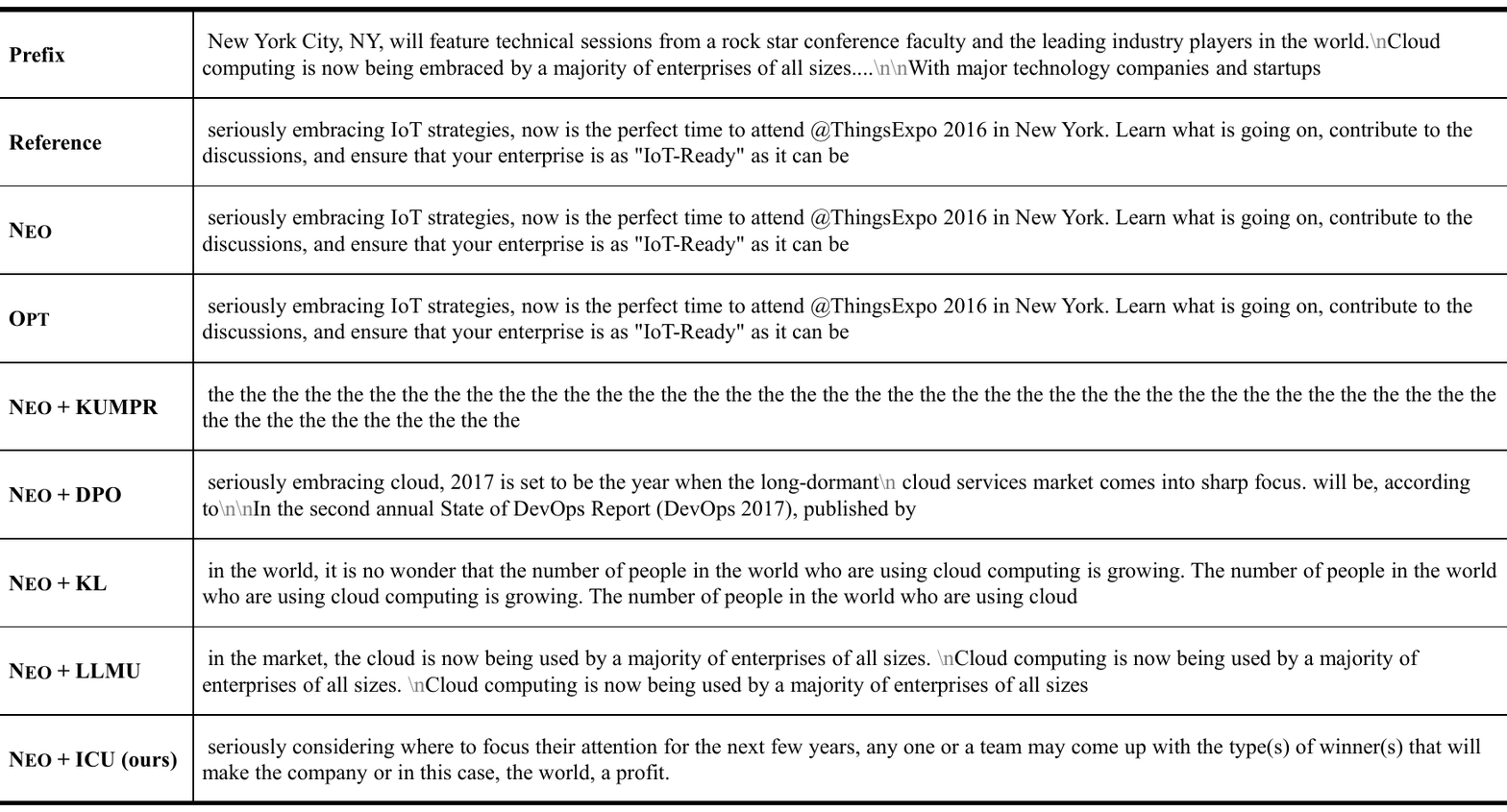}
  \caption{Case study for the comparison among various methods.}
  \vspace{-3mm}
  \label{fig:case}
\end{figure*}

First, the impact of unlearning becomes more pronounced with larger models, indicating that larger models have a higher tendency to memorize sensitive information, which our method effectively mitigates.
Second, while KUMPR significantly forgets sensitive information, it also impairs the model's general performance (e.g., PPL). In contrast, our ICU approach preserves the model's core linguistic capabilities while erasing sensitive information, underscoring its advantage.

\subsection{Ablation study}

As discussed in Section~\ref{sec:CLE}, pair learning loss and KL-divergence loss are employed to ensure the model's stable generative capability. To assess the impact of these losses, Figure~\ref{fig:ablation} shows the performance after removing each loss respectively. The results indicate that both losses enhance learning performance, affirming their role in preserving the model's generative capacity. Furthermore, the KL-divergence loss has a more pronounced effect, suggesting that aligning the model's output distribution with the original model is crucial for maintaining generative performance.

\subsection{Human evaluation}

We have also conducted human evaluation about the results of different methods. Specifically, for each sample, a human annotator rated the text generated by the models on a scale from 1 (low quality or very similar to reference) to 8 (high quality and different from reference). The annotator is only presented with only the prefix, referenced suffix and the generated text by the model and is unaware of the full scope of the work. The instruction and more details about human evaluation can be found in our available codes.

The comparison between GPT and Human Evaluation scores can be found in Table~\ref{tab:annotation}. From the results, human annotation and the GPT evaluation share the same trend, demonstrating the reliability of GPT-4 grading. For statistical test, we normalize the human score $s$ by $10\cdot\frac{s-1}{8-1}$, and the Pearson correlation is $0.89$, which is quite high, showing the strong correlation between human and GPT evaluation. Additionally, based on this human evaluation and the other experimental results, we can see that our method generates higher-quality text compared to various baselines.

\begin{table}[t]
  \caption{Human annotation results on different methods.(Avg. Human scores / Avg. GPT scores.)}
  \label{tab:annotation}
  \centering
  \setlength{\tabcolsep}{3pt} 
  \renewcommand{\arraystretch}{1.1} 
  \begin{tabular}{c|cccc}
    \toprule
    \# params & \textsc{Neo} & \textsc{Opt} & \textsc{Neo} + KUMPR & \textsc{Neo} + ICU (ours) \\
    \midrule
    125M & 3.1 / 3.75 & 3.7 / 3.08 & 1.1 / 1.05 & 5.7 / 3.92 \\
    1.3B & 1.8 / 2.72 & 3.1 / 3.63 & 1.4 / 1.26 & 4.8 / 4.33 \\
    2.7B & 1.5 / 2.11 & 3.2 / 3.65 & 1.1 / 1.00 & 4.7 / 4.40 \\
    
    \bottomrule
  \end{tabular}
  \vspace{-5mm}
\end{table}

\subsection{Case study}

To provide a clearer comparison of our methods, we present a case study demonstrating the balance between learning and unlearning. As shown in Figure~\ref{fig:case}, the reference includes sensitive information such as ``\textit{@ThingsExpo 2016}''. Before unlearning, the original models (e.g., \textsc{GPT-Neo} and \textsc{Opt}) retain and reproduce this information when prompted with the corresponding prefix. When applying the KUMPR method, the models lose their original conversational abilities and repetitively output the word ``\textit{the}''. DPO method also produces sentences not coherent. KL and LLMU methods both repeat same sentence. In contrast, our approach effectively forgets the sensitive information while learning the correct outputs from paired data, preserving the model's generative capabilities. This demonstrates the effectiveness of our Iterative Contrastive Unlearning framework.

\begin{figure*}[t]
  \centering
  \includegraphics[width=0.8\textwidth]{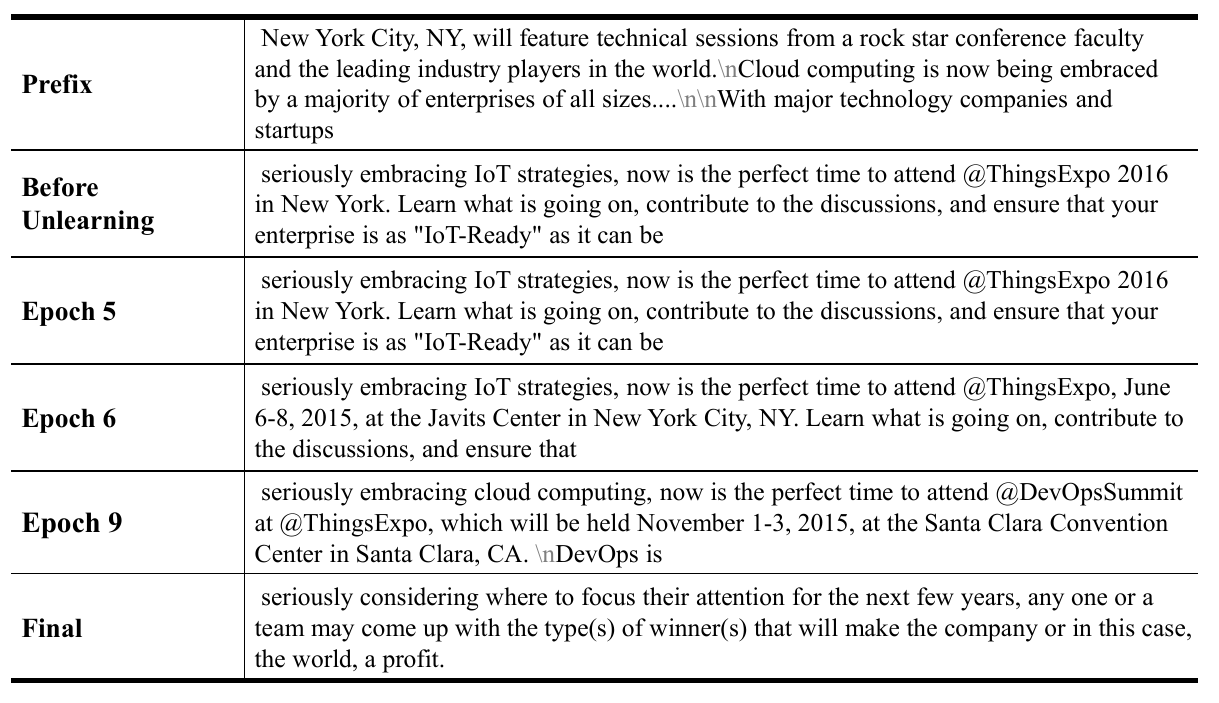}
  \vspace{-3mm}
  \caption{Unlearning process illustration for the case provided in Figure~\ref{fig:case}.}
  \vspace{-3mm}
  \label{fig:effect}
\end{figure*}

\begin{table*}[t]
  \caption{ICU with different $\alpha$ and $\beta$ on \textsc{GPT-Neo} 125M. Our final parameter selection ($\alpha=0.5$ and $\beta=1.0$) is \textbf{bolded}.}
  \label{tab:weight}
  \centering
  \scriptsize
  \setlength{\tabcolsep}{12pt} 
  \renewcommand{\arraystretch}{1} 
  \fontsize{9}{9.5}\selectfont
  \begin{tabular}{cc|ccc|ccc|cc}
    \toprule
    \multirow{2}{*}{\textbf{$\alpha$}} & \multirow{2}{*}{\textbf{$\beta$}} & \textbf{EL}$_{10}$ & \textbf{MA} & \textbf{BERT} & \textbf{Entropy} & \textbf{Cls Avg.} & \textbf{Dia Avg.} & \textbf{Pile} & \textbf{Wikitext}  \\
    & & (\%) $\downarrow$ & (\%) $\downarrow$ & (F1) $\downarrow$ & $\uparrow$ & (ACC) $\uparrow$ & (F1) $\uparrow$ & (PPL) $\downarrow$ & (PPL) $\downarrow$\\
    \midrule
    0.1 & 0.5 & 3.6 & 54.7 & 47.0 & 3.009 & 39.5 & 5.6 & 95.2 & 212.7 \\ 
    0.1 & 1.0 & 4.2 & 56.4 & 52.5 & 3.758 & 42.9 & 10.0 & 23.3 & 44.8 \\
    \midrule
    0.5 & 0.1 & 3.3 & 56.5 & 49.9 & 3.423 & 40.0 & 8.7 & 35.4 & 85.5  \\
    0.5 & 0.5 & 4.2 & 57.5 & 52.6 & 3.755 & 42.4 & 9.7 & 24.5 & 48.8  \\
    \textbf{0.5} & \textbf{1.0} & \textbf{4.4} & \textbf{55.6} & \textbf{53.3} & \textbf{3.833} & \textbf{43.3} & \textbf{10.3} & \textbf{21.6} & \textbf{40.1}  \\ \midrule
    1.0 & 0.5 & 4.5 & 55.1 & 52.1 & 3.700 & 43.1 & 10.1 & 21.6 & 40.1 \\
    1.0 & 1.0 & 4.6 & 53.5 & 52.7 & 3.846 & 43.2 & 10.0 & 21.6 & 40.0 \\
    1.0 & 2.0 & 4.5 & 47.7 & 50.6 & 3.686 & 43.1 & 10.0 & 21.6 & 39.7 \\ \midrule
    2.0 & 1.0 & 4.6 & 47.2 & 50.6 & 3.809 & 42.7 & 9.8 & 22.1 & 40.5  \\
    2.0 & 2.0 & 4.5 & 45.8 & 51.2 & 3.843 & 42.9 & 10.0 & 22.1 & 40.1 \\
    \bottomrule
  \end{tabular}
  \vspace{-4mm}
\end{table*}

\subsection{Unlearning process analysis}\label{app:analysis}
To illustrate the detailed process of unlearning, we provide an example showing the effects at different epochs. As shown in Figure~\ref{fig:effect}, given the prefix, the original model generates texts containing key terms like ``Internet of Things (IoT)'', which are expected to be unlearned. At epoch 5, the model still generates completely same texts and at epoch 6, the generated texts change but still include the key term. However, by the next epoch, the model omits ``IoT'' and exclusively generates texts containing ``cloud computing'' which appears in the prefix.

\label{sec:weight}
\subsection{Parameter sensitivity}
To examine the influence of the loss hyperparameters $\alpha$ and $\beta$ in Section~\ref{sec:CLE}, we conducted extensive parameter sensitivity experiments on \textsc{GPT-Neo} 125M. The results are summarized in Table~\ref{tab:weight}.

\begin{table*}[t]
  \caption{Comparisons on TinyLlama 1.1B. The best comparable performances of unlearning are \textbf{bolded} and second best \underline{underlined}.}
  \label{tab:tl}
  \centering
  \scriptsize
  \setlength{\tabcolsep}{7pt} 
  \renewcommand{\arraystretch}{1.1} 
  \fontsize{8.5}{9}\selectfont
  \begin{tabular}{l|ccc|ccc|cc|c|cc}
    \toprule
    \multirow{2}{*}{\textbf{Model}} & \textbf{EL}$_{10}$ & \textbf{MA} & \textbf{BERT} & \textbf{Entropy} & \textbf{Cls Avg.} & \textbf{Dia Avg.} & \textbf{Pile} & \textbf{Wikitext} & \textbf{GPT} & $\mathbf{s_z}$ & $\mathbf{s_m}$ \\
         & (\%) $\downarrow$ & (\%) $\downarrow$ & (F1) $\downarrow$ & $\uparrow$ & (ACC) $\uparrow$ & (F1) $\uparrow$ & (PPL) $\downarrow$ & (PPL) $\downarrow$ & $\uparrow$ & $\uparrow$ & $\uparrow$ \\
    \midrule
    \textsc{TinyLlama} & 56.2 & 76.8 & 71.2 & 4.742 & 46.2 & 12.4 & 12.8 & 10.7 & - & - & - \\
    \textsc{TinyLlama} + KUMPR & \textbf{0.0} & \textbf{0.3} & \textbf{21.4} & 2.015 & 34.9 & 0.0 & >1000 & >1000 & 1.99 & -3.69 & -1.55 \\
    \textsc{TinyLlama} + DPO & 3.1  & 52.6 & \underline{38.3} & 1.392 & 40.8 & 11.7 & \textbf{13.9} & 11.5 & 1.82 & -3.33 & -1.13 \\
    \textsc{TinyLlama} + KL & 1.8 & 52.2 & 31.9 & 2.753 & 42.8 & \underline{12.3} & 36.7 & 14.0 & 1.63 & 0.37 & 0.20 \\
    \textsc{TinyLlama} + LLMU & 2.5 & 46.0 & 41.9 & \underline{3.683} & 45.6 & \textbf{12.4} & 14.9 & \underline{11.2} & \underline{2.82} & \underline{2.38} & \underline{1.02} \\
    \textbf{\textsc{TinyLlama} + ICU (ours)} & \underline{1.4} & \underline{44.0} & 44.8 & \textbf{3.932} & \textbf{45.8} & \underline{12.3} & \underline{14.3} & \textbf{11.0} & \textbf{3.35} & \textbf{4.28} & \textbf{1.70} \\
    \bottomrule
  \end{tabular}
  \vspace{-5mm}
\end{table*}

In general, increasing $\alpha$ and $\beta$ enhances learning ability while diminishing unlearning ability, with the exception of memorization accuracy (MA). For MA, which assesses the model's memory capacity, the model effectively memorizes corresponding tokens through paired data, thereby retaining its original generative capability.
Meanwhile, we find that when the learning weight $\alpha$ is less than the unlearning weight of 1, variations in the regularization weight $\beta$ significantly impact the model's performance. Conversely, when $\alpha$ exceeds the forgetting weight of 1, changes in $\beta$ do not significantly affect performance. This indicates that both hyperparameters contribute to increased learning ability, corroborating the findings presented in Figure~\ref{fig:ablation}.
To balance the model's learning and forgetting abilities, we ultimately selected \(\alpha = 0.5\) and \(\beta = 1.0\) as our reported parameters.

\subsection{Generalization ability of ICU}\label{app:generalization}

In this section, we discuss the generalization ability of our proposed Iterative Contrastive Unlearning framework to GLMs other than the \textsc{GPT-Neo} model. 
As introduced in Section~\ref{sec:ICU_framework}, our proposed ICU framework for generative LMs, which can be applied to various advanced GLMs, such as Llama, Bloom, etc.
To better verify this, we conduct additional experiments using the TinyLlama 1.1B model~\cite{zhang2024tinyllama}, which features a different architecture from the \textsc{GPT-Neo} model in the main experiment. Specifically, we compared our ICU method with KUMPR~\cite{jang2023knowledge}, DPO~\cite{maini2024tofu}, KL~\cite{maini2024tofu}, and LLMU~\cite{yao2023large}. The results are summarized in Table~\ref{tab:tl}. With the best normalized scores, our method surpasses all the methods in terms of balancing the unlearning and preserving model abilities, demonstrating the superior generalization ability of ICU framework.

\section{Conclusion}

In this work, we explored machine unlearning for Generative Language Models, focusing on the challenge of selectively removing sensitive data while preserving overall model performance. We proposed the Iterative Contrastive Unlearning (ICU) framework. Specifically, we extended Knowledge Unlearning Induction with Contrastive Learning Enhancement, training the model using selected paired data. Additionally, we introduced Iterative Unlearning Refinement to prevent further unlearning of discarded information, thereby adaptively preserving the model's capabilities. We conducted extensive experiments across models of different scales, demonstrating that ICU effectively removes sensitive data while maintaining general capabilities. Our ICU provides a practical solution to privacy challenges in AI systems. Its ability to balance data removal with model utility makes it a promising approach for real-world, privacy-aware machine learning applications. Future work will focus on exploring such domains, where safeguarding sensitive information without compromising general ability is essential.

\section{Acknowledgements}
This research was partially supported by the National Natural Science Foundation of China (U23A20319, 62441239,2406303), Anhui Provincial Natural Science Foundation (No. 2308085QF229), Anhui Province Science and Technology Innovation Project (202423k09020010), the Fundamental Research Funds for the Central Universities (No. WK2150110034).

\bibliographystyle{IEEEtran}
\bibliography{IEEEabrv,IEEEexample}

\end{document}